\documentclass{article}
\usepackage{spconf,amsmath,graphicx}
\usepackage{booktabs}

\usepackage{spconf,amsmath,graphicx}
\usepackage{amsfonts}
\usepackage{amssymb}
\usepackage{url}
\usepackage{float}

\usepackage{times}
\usepackage{latexsym}
\usepackage{hyperref}
\usepackage{url}
\usepackage{amsfonts}
\usepackage{graphicx}
\usepackage{amsmath}
\usepackage{booktabs}
\usepackage{multirow}
\usepackage{bbm}
\usepackage{float}

\title{Sequence-Level Knowledge Distillation for Model Compression of Attention-based Sequence-to-Sequence Speech Recognition}
\name{Raden Mu'az Mun'im, Nakamasa Inoue, Koichi Shinoda}
\address{
Tokyo Institute of Technology\\
raden@ks.c.titech.ac.jp, inoue@ks.c.titech.ac.jp, shinoda@c.titech.ac.jp
}

\begin{document}
\newcommand{\fix}{\marginpar{FIX}}
\newcommand{\new}{\marginpar{NEW}}
\newcommand{\xvec}{\mathbf{x}}
\newcommand{\yvec}{\mathbf{y}}
\newcommand{\cvec}{\mathbf{c}}
\newcommand{\zvec}{\mathbf{z}}
\newcommand{\svec}{\mathbf{s}}
\newcommand{\tvec}{\mathbf{t}}
\newcommand{\mcL}{\mathcal{L}}
\newcommand{\mcT}{\mathcal{T}}
\newcommand{\mcY}{\mathcal{Y}}
\newcommand{\mcV}{\mathcal{V}}
\newcommand{\mcC}{\mathcal{C}}
\newcommand{\mcA}{\mathcal{A}}
\newcommand{\context}{\mathbf{y}_{\mathrm{c}}}
\newcommand{\embcontext}{\mathbf{\tilde{y}}_{\mathrm{c}}}
\newcommand{\inpcontext}{\mathbf{\tilde{x}}}
\newcommand{\start}{\mathbf{\tilde{y}}_{\mathrm{c0}}}
\newcommand{\End}{\mathrm{\texttt{</s>}}}

\newcommand{\Uvec}{\mathbf{U}}
\newcommand{\Evec}{\mathbf{E}}
\newcommand{\Gvec}{\mathbf{G}}
\newcommand{\Fvec}{\mathbf{F}}
\newcommand{\Pvec}{\mathbf{P}}
\newcommand{\pvec}{\mathbf{p}}
\newcommand{\Vvec}{\mathbf{V}}
\newcommand{\Wvec}{\mathbf{W}}
\newcommand{\hvec}{\mathbf{h}}
\newcommand{\wvec}{\mathbf{w}}
\newcommand{\uvec}{\mathbf{u}}
\newcommand{\vvec}{\mathbf{v}}
\newcommand{\bvec}{\mathbf{b}}
\newcommand{\reals}{\mathbb{R}}
\newcommand\given{\,|\,}
%
\maketitle
\begin{abstract}
We investigate the feasibility of sequence-level knowledge distillation of Sequence-to-Sequence (Seq2Seq) models for  Large Vocabulary Continuous Speech Recognition (LVSCR). We first use a pre-trained larger teacher model to generate multiple hypotheses per utterance with beam search. With the same input, we then train the student model using these hypotheses generated from the teacher as pseudo labels in place of the original ground truth labels. We evaluate our proposed method using Wall Street Journal (WSJ) corpus. It achieved up to $ 9.8 \times$ parameter reduction with accuracy loss of up to 7.0\% word-error rate (WER) increase.

\end{abstract}
\section{Introduction}
\label{sec:intro}
In recent years, end-to-end deep neural networks for Large Vocabulary Continuous Speech Recognition (LVSCR) have been steadily improving their accuracy, rivalling the traditional Gaussian Mixture Model-Hidden Markov Models (GMM-HMM) and hybrid models of deep neural networks and HMMs (DNN-HMM). While these models have an acoustic model and a language model which are trained separately, for end-to-end training, the whole model is trained using backpropagation with audio-transcription pairs \cite{sutskever2014sequence}. 

Several end-to-end architectures for speech recognition have been proposed. Their examples include Connectionist Temporal Classification (CTC) \cite{graves2013speech}, Recurrent Neural Network-Transducer (RNN-T) \cite{graves2014towards}, and Sequence-to-Sequence (Seq2Seq) with attention 
 \cite{bahdanau2016end}. Unlike CTC and RNN-T, Seq2Seq with attention does not make any prior assumptions on the output sequence alignment given an input; it jointly learns how to align while learning to encode the input and decode its result into the output. 

In machine learning, model compression \cite{han2015deep} is a way to significantly compress a model by reducing the number of its parameters while having negligible accuracy loss. This is important for deploying trained models on memory and compute-constrained devices such as mobile phones and embedded systems, and when energy efficiency is needed on large-scale deployment. Several model compression methods exist, such as pruning, quantization \cite{han2015deep}, and knowledge distillation (KD) \cite{hinton2015distilling}. KD is the focus of this work, where a smaller student model is trained by using the output distribution of a larger teacher model.   Recently KD for Seq2Seq models with attention was proposed for neural machine translation (NMT) task \cite{kim2016sequence} and proved to be effective. Also, Seq2Seq models for CTC-based speech recognition was recently proposed \cite{takashima2018ctc}.

 In this paper, we propose a sequence-level KD method for Seq2Seq speech recognition models with attention. Different from the previous work for NMT \cite{kim2016sequence}, we extract the hypotheses from a pre-trained teacher Seq2Seq model using beam search, and train student Seq2Seq models using the hypotheses as pseudo labels on the sequence-level cross-entropy criterion. To the best of our knowledge, there is no prior work on applying KD in Seq2Seq-based speech recognition models.


\section{Knowledge Distillation for Seq2Seq Models}

\subsection{Sequence-to-Sequence Learning}

The Sequence-to-Sequence (Seq2Seq) \cite{sutskever2014sequence} is neural network architecture which directly models conditional probability $p(\mathbf{y}|\mathbf{x})$ where $\mathbf{x} = [x_1, ..., x_S]$ is the source sequence with length $S$ and $\mathbf{y} = [y_1, ..., y_T]$ is the target sequence with length $T$.

\begin{figure}[t!]
	\centering
	\includegraphics[width=0.3\linewidth]{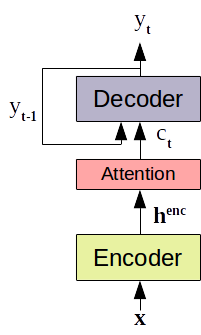}
	\caption{Architecture of Seq2Seq with Attention module}
	\label{fig:seq2seq}
\end{figure}

Figure \ref{fig:seq2seq} illustrates the Seq2Seq model. It consists of an encoder, a decoder and an attention module. 
The encoder processes an input sequence $\mathbf{x}$ and outputs encoded hidden representation $\mathbf{h^e} = [h^e_1, ...,h^e_S]$ for the decoder \cite{tjandra2017attention}. The attention module is a network that assists the decoder to find relevant information on the encoder side based on the current decoder hidden states \cite{bahdanau2016end}. The attention module does this by producing a context vector $c_t$ at time $t$ based on the encoder and decoder hidden states:
\begin{align}
c_t &= \sum_{s=1}^{S} a_t(s) * h^e_s , \\
a_t(s) & = \frac{\exp(\text{Score}(h^e_s, h^d_t))}{\sum_{s=1}^{S}\exp(\text{Score}(h^e_s, h^d_t))} ,
\end{align}
Several variations exist for $\text{Score}(h^e_s, h^d_t)$:
\begin{align}
\text{Score}(h_s^e, h_t^d) =
\begin{cases}
\langle h_s^e, h_t^d\rangle ,& \text{: dot product}  \\
h_s^{e\intercal} W_{s} h_t^d ,& \text{: bilinear}  \\
V_s^{\intercal} \tanh(W_{s} [h_s^e, h_t^d]) ,& \text{: MLP} \label{eq:mlpscore}  \\
\end{cases}
\end{align} where $\text{Score}$ is a function $(\mathbb{R}^M \times \mathbb{R}^N) \rightarrow \mathbb{R}$, $M$ is the number of hidden units for the encoder, $N$ is the number of hidden units for the decoder, and both $W_s$ and $V_s^{\intercal}$ are weight matrices.
Finally, the decoder predicts the probability of target sequence $\mathbf{y}$ at time $t$ based on the previous output $y_{<t}$ and $c_t$, which can be formulated as:

\begin{equation}
\log{p(\mathbf{y}|\mathbf{x})} = \sum_{t=1}^{T}\log{p(y_t|y_{<t}, c_t)} .
\end{equation}

 The previous outputs $c_t$ can obtained with greedy decoding, i.e. by taking the output with the highest probability for each time step (e.g in \cite{kim2016sequence}). It is also possible to perform \textit{beam search} to obtain more than one $c_t$ \cite{sutskever2014sequence}.

Seq2Seq can handle virtually any sequence related tasks\cite{sutskever2014sequence}, such as NMT and speech recognition. For speech recognition, the input $\mathbf{x}$ is a sequence of feature vectors derived from audio waveform such as Short-time Fourier Transform (STFT) spectrogram or  Mel-frequency cepstral coefficients (MFCC). Therefore, $ \mathbf{x} $ is a real vector with ${S \times D}$ dimension where D is the number of the features and S is the total length of the utterance in frames. The output $\mathbf{y}$ can be a sequence of phonemes or graphemes (characters).

\subsection{Knowledge Distillation}
\label{sec:pagestyle}

In knowledge distillation (KD), a teacher model's probability distribution  $q(\theta_T | x; \theta)$ is trained by using a given dataset, where $\theta_t$ is a set of its parameters. Using the same dataset, a student model $p(\theta | x; \theta)$ is trained by minimizing cross-entropy with the teacher model's probability distribution  $q(\theta_T | x; \theta)$ instead of the ground truth labels from the dataset.

Let an input-label pairs be $(x, y)$ and $\mcV$ is a set of possible classes. Then, the loss for KD is given as:

\begin{equation}
\mcL_{\text{KD}}(\theta;\theta_T) = - \sum_{k=1}^{|\mcV|}  q(y = k \given x;\theta_T)\log p(y = k \given x;\theta) .
\end{equation}

In KD training, the teacher produces softmax probabilities (\textit{soft targets}), which reveal teacher's confidences for what classes it predicts given an input. With this additional knowledge, the student can model the data distribution better than learning directly from the ground truth labels consisting of one-hot vectors (\textit{hard targets})  \cite{hinton2015distilling}.


\subsection{Sequence-Level Knowledge Distillation}
While it is possible to use the original KD method to train autoregressive models such as RNN, it only gives non-significant accuracy gains \cite{kim2016sequence}, or simply degrade the performance  compared to training with the dataset directly \cite{sak2015acoustic}.

To adapt KD to autoregressive models, \cite{kim2016sequence} suggested to use the approximation of a teacher's sequence-level distribution instead of the teacher's single time step frame-level distribution, 
so as to capture the time-dependencies between the inputs. 

Consider the sequence-level distribution
specified by the model over all possible teacher label sequences $\tvec \in \mcT$, given the input sequence \textbf{s} :
\begin{equation}
p(\tvec \given \svec) = \prod_{j=1}^J p(t_j \given \svec, \tvec_{<j}),
\end{equation}
for any length $J$.
The sequence-level loss for Seq2Seq involves matching
the input \textbf{s} with the one-hot distribution of  all the complete sequences:
\begin{eqnarray}
	&&\mcL_{\text{SEQ-NLL}} = -\sum_{\tvec \in \mcT} \mathbbm{1}\{\tvec=\yvec\} \log p(\tvec \given \svec) \\ 
	&= & -\sum_{j=1}^J \sum_{k=1}^{|\mcV|} \mathbbm{1}\{y_j=k\} \log p(t_{j}=k \given \svec, \tvec_{<j}), 	 
\end{eqnarray}
where $\mathbbm{1}\{\cdot\}$ is the indicator function and $\yvec = [y_1, \dots, y_J]$ is the observed sequence. 

To formulate the sequence-level KD, we use $q(\tvec \given \svec)$ to represent
the teacher's sequence distribution over the sample space of
all possible sequences,
\begin{equation}
\mcL_{\text{SEQ-KD}} = -\sum_{\tvec \in \mcT} q(\tvec \given \svec) \log p (\tvec \given \svec)
\end{equation}

Different from previously stated $\mcL_\text{KD}$, $\mcL_\text{SEQ-KD}$ minimizes the loss on the whole-sequence level.   However, this loss is intractable. An approximation to calculate
it is required. 

\begin{figure}[t!]
	\centering
	\includegraphics[width=1
	\linewidth]{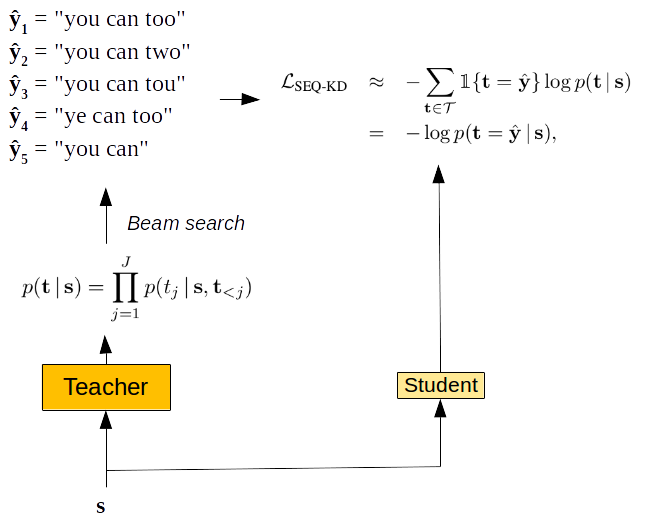}
	\caption{Example of sequence-level KD training. The teacher produces pseudo labels obtained from the top-\textit{k} results from beam search, then the student learns by minimizing cross entropy with them.}
	\label{fig:seq-kd}
\end{figure}

There are many ways to approximate the loss. The sequence-level KD for NMT \cite{kim2016sequence} uses a single hypothesis with the best score as the pseudo label per input.
For CTC-based speech recognition \cite{takashima2018ctc}, this loss was \textit{k}-best hypotheses from beam search per input (Figure \ref{fig:seq-kd}).
The loss is then
\begin{eqnarray}
	\mcL_{\text{SEQ-KD}} &\approx&  - \sum_{\tvec \in \mcT} \mathbbm{1}\{\tvec = \hat{\yvec} \} \log p (\tvec \given \svec) \\
	&=& - \log p (\tvec = \hat{\yvec} \given \svec),
\end{eqnarray}
where 
$\hat{\yvec}$ is the output hypothesis obtained from running beam search with the teacher model. In this work, we investigated on how to apply these for Seq2Seq-based speech recognition models, which will be discussed in the next sections.

\section{Multiple Hypotheses from Beam search for Knowledge Distillation}

For Seq2Seq-based speech recognition, the sequence-level KD with the loss approximation can be done by the following three steps: (1) Pre-train a teacher model with a dataset, (2) With the same dataset, generate \textit{k}-best sequences with beam search, and save them as pseudo labels, (3) Train a student model with the same dataset but replace the ground truth labels with generated pseudo labels (Figure \ref{fig:seq-kd}). In this procedure, the size of beam search and the value of \textit{k} are adjustable hyperparameters. The dataset size increases with factor of \textit{k} from the method using the 1-best \cite{kim2016sequence}.

 The pseudo labels are analogous to \textit{soft targets} in the original KD. 
 Even if the pseudo labels are not fully accurate, the student is expected to achieve better performance with this training method since the student tries to imitate the teacher's distribution instead of trying to model the training data distribution directly. Training with these pseudo labels can be seen as a form of regularization similar to the original KD \cite{hinton2015distilling}, because the slightly inaccurate transcriptions from pseudo labels can prevent the trained models from overfitting to training data distribution (Figure \ref{fig:seq-kd}).

\section{Implementation and Experiments}

\subsection{Seq2Seq Model Architecture}
The input speech audio waveform is sampled at 16kHz, then transformed into STFT spectrogram with Hanning window of 20ms and step size of 10ms. Then the STFT spectrograms are fed into 2D convolutional neural network (CNN) with two layers, as described by \cite{maas2017building, amodei2016deep}, as this STFT and CNN combination can further improve accuracy compared to using MFCC alone. 2D CNN is configured with filters=32, kernel\textunderscore size=(5, 8), stride=(2,2).

Bidirectional Gated Recurrent Unit (GRU) \cite{chung2014empirical} is used for the encoder and the decoder. The feature vector from 2D CNN is first fed into the encoder. Then, the hidden representation made by the encoder is fed to an embedding layer of size 32. Next, the output of the embedding layer is fed into the decoder.

The output of the decoder is fed into the softmax layer consisting of 31 classes (26 English alphabet plus 5 classes for start-of-sentence (sos), end-of-sentence (eos),  space, apostrophe and period). The decoder's attention module consists of 1D CNN layer with kernel=128, kernel\textunderscore size=15, stride=1, padding=7 followed by a fully-connected layer, as proposed by \cite{bahdanau2016end}. The models are trained with Dropout set to 0.4.

The model configurations are summarised in Table 1. We have designed two kinds of student models with different sizes, Student-mid and Student-small.

\begin{table}[t!]
	\label{tab:arch}
	\begin{tabular}{l l l l l}
		\hline
		Model &  Encoder bi-GRU & Decoder bi-GRU \\
		\hline
		
		Teacher & 5 layers 384 cells & 3 layers 384 cells \\
		Student-mid & 4 layers 256 cells & 1 layer 256 cells \\
		Student-small & 3 layers 128 cells & 1 layer 128 cells \\
		\hline
		\hline

	\end{tabular}
	\caption{Model configurations. 2D CNN and Attention module configuration is the same for all models.}%
\end{table}

\subsection{Dataset and Software}
Wall Street Journal (WSJ) corpus (available at the Linguistic Data Consortium
as LDC93S6B and LDC94S13B) is used as the training and testing datasets. From the corpus, train\textunderscore si284 (81 hours of 37k sentences) is used for training, dev93 is used for validation, and eval92 is used for testing. Data is extracted as according to WSJ recipe from Kaldi toolkit \cite{povey2011kaldi}. STFT spectrogram extraction, implementation of the models, training and testing are conducted using Python 3.6, Scipy 1.0.1 and Pytorch 0.3.1.
\subsection{Experimental Setup}
First, the teacher model and student models are trained directly with train\textunderscore si284 and tested with eval92 to serve as the baselines. To perform Sequence-level KD, the teacher model pre-trained with train\textunderscore si284 is used for pseudo labels generation. This is done by extracting \textit{k}-best hypotheses from beam search with combinations of beam size of 5, 10 and \textit{k}-best of 1, 5, 10. 

For each model the training is done using Adam optimizer, with learning rate of 2e-4 exponential decay rate of 0.99 per epoch, and mini-batch size set at 16. Seq2Seq teacher forcing rate \cite{bahdanau2016end} is set at 0.4.  Training is done up to 200 epochs, or until no improvement can be seen in validation or testing set.

%
%
\begin{table}[tb!]
	\centering
	\label{tab:results}
	\caption{Results on WSJ “eval92” (trained on “train\_si284”). beamSize (shorthand for size of beam search) and topK (shorthand for top \textit{k}-best hypotheses) are hyperparameters. We measured the accuracy using the character-error rate (CER) and word-error rate (WER).}%
	\begin{tabular}{l c c c c }
		Model &CER (\%) &WER (\%) &\\
		\hline
		Teacher \\(Params: 16.8M; $ 100 \%$ size)\\
		\hspace{1.0cm} Baseline &4.6  &15.3 \\
		\hline
		Student-mid \\(Params: 6.1M; $ 37 \%$ size)\\
		\hspace{1.0cm} Baseline &7.0  &21.8\\
		\hspace{1.0cm} topK=1, beamSize=5 	&6.4  &20.5 \\
		\hspace{1.0cm} topK=1, beamSize=10 	&6.5  &20.5 \\
		\hspace{1.0cm} topK=5, beamSize=5 	&\textbf{6.0}  &20.1 \\
		
		\hspace{1.0cm} topK=5, beamSize=10	&6.5  &21.2 \\ 
		\hspace{1.0cm} topK=10, beamSize=10 	&6.1  &\textbf{19.7} \\ 

		\hline
		Student-small \\(Params: 1.7M; $ 10 \%$ size)\\
		
		\hspace{1.0cm} Baseline 	&9.2  &28.7 \\
		\hspace{1.0cm} topK=1, beamSize=5 	&7.6  &26.1 \\
		\hspace{1.0cm} topK=1, beamSize=10 &7.7 &25.3 \\ 
		\hspace{1.0cm} topK=5, beamSize=5 &\textbf{6.5}  &\textbf{22.3} \\ 
		
		\hspace{1.0cm} topK=5, beamSize=10 &6.9  &23.3 \\ 
				\hspace{1.0cm} topK=10, beamSize=10 	&7.4  &24.7 \\

		\hline 
		\hline
	
	\end{tabular}
\end{table}
\section{Results}
The results are shown in Table 2. 
The teacher model serves the reference baseline for student models. 
We benchmarked the character-error rate (CER) and word-error rate (WER). With KD training, the student models managed to achieve better CER and WER than the case when training directly with the dataset.




For Student-mid model, the number of parameters is $37\%$ of the teacher ($2.7\times$ reduction). It achieved the best CER (6.0\%) with beamSize=5 and topK=5, and the best WER (19.7\%) with beamSize=10 and beamSize=10.

For Student-small model, the number of parameters is $10\%$ of the teacher ($9.8\times$ reduction). As expected, the model suffered higher CER and WER compared to Student-mid model. The model achieved the best CER (6.5\%) and WER (22.3\%) with beamSize=5 and topK=5. The effect of sequence-level KD is more obvious in Student-small, where it achieve 6.4\% reduction in WER with KD training compared to directly training with the dataset.

Training was done using a server with Intel Xeon E5-2680 2.4GHz and NVIDIA Tesla P100 16GB. We did not found significant improvement in training and testing time. Student-mid and Student-small achieved speed-up of $1.4\times$ and $1.7\times$ respectively, relative to the teacher model. This relatively small improvement may be due to inherent sequential computations using RNN where some operations simply cannot be paralellized.

To summarise, generally we found that using beamSize=5 and topK=5 are sufficient for reasonable WER and CER reduction. Increasing beamSize and/or topK further do not necessarily improve the performance.
%
%
\section{Conclusion}

In this work, we successfully performed sequence-level KD training for Seq2Seq speech recognition models. Using beam search to approximate the teacher's distribution, we extracted \textit{k}-best hypotheses to be used as pseudo labels to train the student on the same dataset. We managed to train the students with reduction of $9.8\times$ parameter size of the teacher, with increase of WER of $7.0\%$ relative to the teacher. 

There are many problems left for future work. Since RNN is limited in inference speed due to its sequential operations, we plan to investigate the feasibility of sequence-level KD for other highly parallelizable attention-based architectures which are not based on RNN, such as, Transformer \cite{vaswani2017attention} and S2SConv \cite{gehring2017convolutional}.

\section{Acknowledgment}
This work was supported by JST CREST Grant Number JPMJCR1687 and JSPS
KAKEN Grant Number 16H02845.


%

\pagebreak


\bibliographystyle{IEEEbib}
\bibliography{refs}

\begin{thebibliography}{10}

\bibitem{sutskever2014sequence}
Ilya Sutskever, Oriol Vinyals, and Quoc~V Le,
\newblock ``Sequence to sequence learning with neural networks,''
\newblock in {\em Advances in neural information processing systems}, 2014, pp.
  3104--3112.

\bibitem{graves2013speech}
Alex Graves, Abdel-rahman Mohamed, and Geoffrey Hinton,
\newblock ``Speech recognition with deep recurrent neural networks,''
\newblock in {\em Acoustics, speech and signal processing (icassp), 2013 IEEE
  international conference on}. IEEE, 2013, pp. 6645--6649.

\bibitem{graves2014towards}
Alex Graves and Navdeep Jaitly,
\newblock ``Towards end-to-end speech recognition with recurrent neural
  networks,''
\newblock in {\em International Conference on Machine Learning}, 2014, pp.
  1764--1772.

\bibitem{bahdanau2016end}
Dzmitry Bahdanau, Jan Chorowski, Dmitriy Serdyuk, Philemon Brakel, and Yoshua
  Bengio,
\newblock ``End-to-end attention-based large vocabulary speech recognition,''
\newblock in {\em Acoustics, Speech and Signal Processing (ICASSP), 2016 IEEE
  International Conference on}. IEEE, 2016, pp. 4945--4949.

\bibitem{han2015deep}
Song Han, Huizi Mao, and William~J Dally,
\newblock ``Deep compression: Compressing deep neural networks with pruning,
  trained quantization and huffman coding,''
\newblock {\em arXiv preprint arXiv:1510.00149}, 2015.

\bibitem{hinton2015distilling}
Geoffrey Hinton, Oriol Vinyals, and Jeff Dean,
\newblock ``Distilling the knowledge in a neural network,''
\newblock {\em arXiv preprint arXiv:1503.02531}, 2015.

\bibitem{kim2016sequence}
Yoon Kim and Alexander~M Rush,
\newblock ``Sequence-level knowledge distillation,''
\newblock in {\em Proceedings of the 2016 Conference on Empirical Methods in
  Natural Language Processing}, 2016, pp. 1317--1327.

\bibitem{chorowski2015attention}
Jan~K Chorowski, Dzmitry Bahdanau, Dmitriy Serdyuk, Kyunghyun Cho, and Yoshua
  Bengio,
\newblock ``Attention-based models for speech recognition,''
\newblock in {\em Advances in neural information processing systems}, 2015, pp.
  577--585.

\bibitem{tjandra2017attention}
Andros Tjandra, Sakriani Sakti, and Satoshi Nakamura,
\newblock ``Attention-based wav2text with feature transfer learning,''
\newblock in {\em Automatic Speech Recognition and Understanding Workshop
  (ASRU), 2017 IEEE}. IEEE, 2017, pp. 309--315.

\bibitem{sak2015acoustic}
Ha{\c{s}}im Sak, F{\'e}lix de~Chaumont~Quitry, Tara Sainath, Kanishka Rao,
  et~al.,
\newblock ``Acoustic modelling with cd-ctc-smbr lstm rnns,''
\newblock in {\em Automatic Speech Recognition and Understanding (ASRU), 2015
  IEEE Workshop on}. IEEE, 2015, pp. 604--609.

\bibitem{takashima2018ctc}
Ryoichi Takashima, Sheng Li, and Hisashi Kawai,
\newblock ``An investigation of a knowledge distillation method for ctc
  acoustic models,''
\newblock in {\em Acoustics, Speech and Signal Processing (ICASSP), 2018 IEEE
  International Conference on}. IEEE, 2018, pp. 5809--5813.

\bibitem{maas2017building}
Andrew~L Maas, Peng Qi, Ziang Xie, Awni~Y Hannun, Christopher~T Lengerich,
  Daniel Jurafsky, and Andrew~Y Ng,
\newblock ``Building dnn acoustic models for large vocabulary speech
  recognition,''
\newblock {\em Computer Speech \& Language}, vol. 41, pp. 195--213, 2017.

\bibitem{amodei2016deep}
Dario Amodei, Sundaram Ananthanarayanan, Rishita Anubhai, Jingliang Bai, Eric
  Battenberg, Carl Case, Jared Casper, Bryan Catanzaro, Qiang Cheng, Guoliang
  Chen, et~al.,
\newblock ``Deep speech 2: End-to-end speech recognition in english and
  mandarin,''
\newblock in {\em International Conference on Machine Learning}, 2016, pp.
  173--182.

\bibitem{chung2014empirical}
Junyoung Chung, Caglar Gulcehre, Kyunghyun Cho, and Yoshua Bengio,
\newblock ``Empirical evaluation of gated recurrent neural networks on sequence
  modeling,''
\newblock in {\em NIPS 2014 Workshop on Deep Learning, December 2014}, 2014.

\bibitem{povey2011kaldi}
Daniel Povey, Arnab Ghoshal, Gilles Boulianne, Lukas Burget, Ondrej Glembek,
  Nagendra Goel, Mirko Hannemann, Petr Motlicek, Yanmin Qian, Petr Schwarz,
  et~al.,
\newblock ``The kaldi speech recognition toolkit,''
\newblock in {\em IEEE 2011 workshop on automatic speech recognition and
  understanding}. IEEE Signal Processing Society, 2011, number
  EPFL-CONF-192584.

\bibitem{vaswani2017attention}
Ashish Vaswani, Noam Shazeer, Niki Parmar, Jakob Uszkoreit, Llion Jones,
  Aidan~N Gomez, {\L}ukasz Kaiser, and Illia Polosukhin,
\newblock ``Attention is all you need,''
\newblock in {\em Advances in Neural Information Processing Systems}, 2017, pp.
  5998--6008.

\bibitem{gehring2017convolutional}
Jonas Gehring, Michael Auli, David Grangier, Denis Yarats, and Yann~N Dauphin,
\newblock ``Convolutional sequence to sequence learning,''
\newblock in {\em International Conference on Machine Learning}, 2017, pp.
  1243--1252.

\end{thebibliography}

\end{document}